\documentclass[nohyperref]{article}

\usepackage{microtype}
\usepackage{graphicx}
\usepackage{subfigure}
\usepackage{booktabs} % for professional tables

\usepackage{hyperref}

\usepackage[accepted]{icml2023}

\usepackage{amsmath}
\usepackage{amssymb}
\usepackage{mathtools}
\usepackage{amsthm}

\usepackage[capitalize,noabbrev]{cleveref}

\theoremstyle{plain}

\theoremstyle{definition}

\theoremstyle{remark}

\usepackage[textsize=tiny]{todonotes}
\usepackage{multirow}
\usepackage[misc]{ifsym}

\icmltitlerunning{RCsearcher: Reaction Center Identification in Retrosynthesis via Deep Q-Learning}

\begin{document}

\twocolumn[
\icmltitle{RCsearcher: Reaction Center Identification in Retrosynthesis \\ via Deep Q-Learning}



\icmlsetsymbol{equal}{*}

\begin{icmlauthorlist}
\icmlauthor{Zixun Lan}{yyy}
\icmlauthor{Zuo Zeng}{comp}
\icmlauthor{Binjie Hong}{yy}
\icmlauthor{Zhenfu Liu}{comp}
\icmlauthor{Fei Ma$^{\textrm{\Letter}}$}{yyy}
\end{icmlauthorlist}

\icmlaffiliation{yyy}{Department of Applied Mathematics, Xi’an Jiaotong-Liverpool University, Suzhou, China}
\icmlaffiliation{comp}{Hours Technology Co., Ltd., Suzhou, China}
\icmlaffiliation{yy}{Department of Information and Computing Science, Xi’an Jiaotong-Liverpool University, Suzhou, China}

\icmlcorrespondingauthor{Fei ma}{Fei.Ma@xjtlu.edu.cn}

\icmlkeywords{Machine Learning, ICML}

\vskip 0.3in
]



\printAffiliationsAndNotice{} % otherwise use the standard text.

\begin{abstract}
The reaction center consists of atoms in the product whose local properties are not identical to the corresponding atoms in the reactants. Reaction center identification plays a vital role in single-step retrosynthesis. Prior studies on reaction center identification mainly appear in semi-templated retrosynthesis methods. Moreover, they are limited to single reaction center identification. However, many reaction centers are comprised of multiple bonds or atoms in reality. We refer to this kind of reaction center as the multiple reaction center. This paper presents RCsearcher, a unified framework for single and multiple reaction center identification that combines the advantages of the graph neural network and deep reinforcement learning. The critical insight in this framework is that the single or multiple reaction center must be a node-induced subgraph of the molecular product graph. Comprehensive experiments demonstrate that RCsearcher consistently outperforms other baselines for the reaction centre identification task and can extrapolate the reaction center patterns that have not appeared in the training set. Ablation experiments verify the effectiveness of individual components, including the beam search and one-hop-constraint of action space.
\end{abstract}

\section{Intruduction}
Reaction center identification plays a vital role in single-step retrosynthesis. The reaction center (RC) consists of atoms in the product whose local properties are not identical to the corresponding atoms in the reactants \cite{coley2019rdchiral}. Experienced chemists first disconnect the target molecule through the potential reaction center when performing retrosynthesis analysis \cite{corey1991logic}. In computer-aided retrosynthesis, template-based single-step retrosynthesis methods \cite{coley2017computer,dai2019retrosynthesis,chen2021deep} rank the reaction templates, where reaction centers are part of the reaction template. The first step of the semi-template-based retrosynthesis methods \cite{shi2020graph,yan2020retroxpert,somnath2021learning} is to use one graph neural network to predict the reaction center of the target molecule. Moreover, some template-free retrosynthesis methods \cite{wan2022retroformer,wang2021retroprime} began explicitly or implicitly adding reaction center information to improve performance and interpretability.

\begin{figure}[t]
	\centering
	\includegraphics[width=0.85\linewidth]{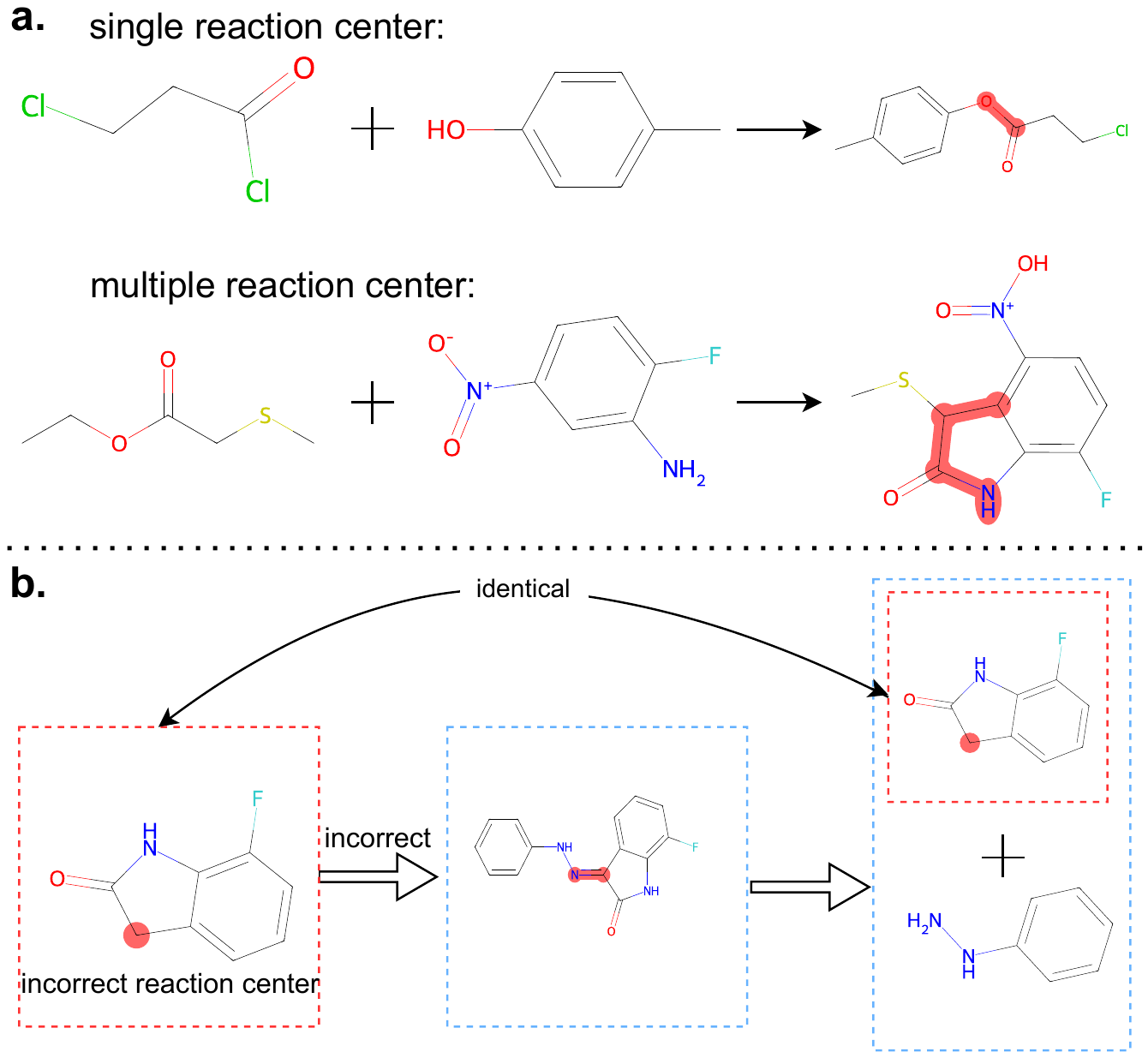}
	\caption{(a) Examples of Single reaction center and multiple reaction center. The red highlight is the reaction center. (b) The case of the same intermediate molecules are visited many times in the multi-step retrosynthesis. The red highlight and arrow denote the predicted reaction center and retrosynthetic prediction respectively.}
	\label{intro1}
\end{figure}

\begin{figure*}[t]
	\centering
	\includegraphics[width=0.95\linewidth]{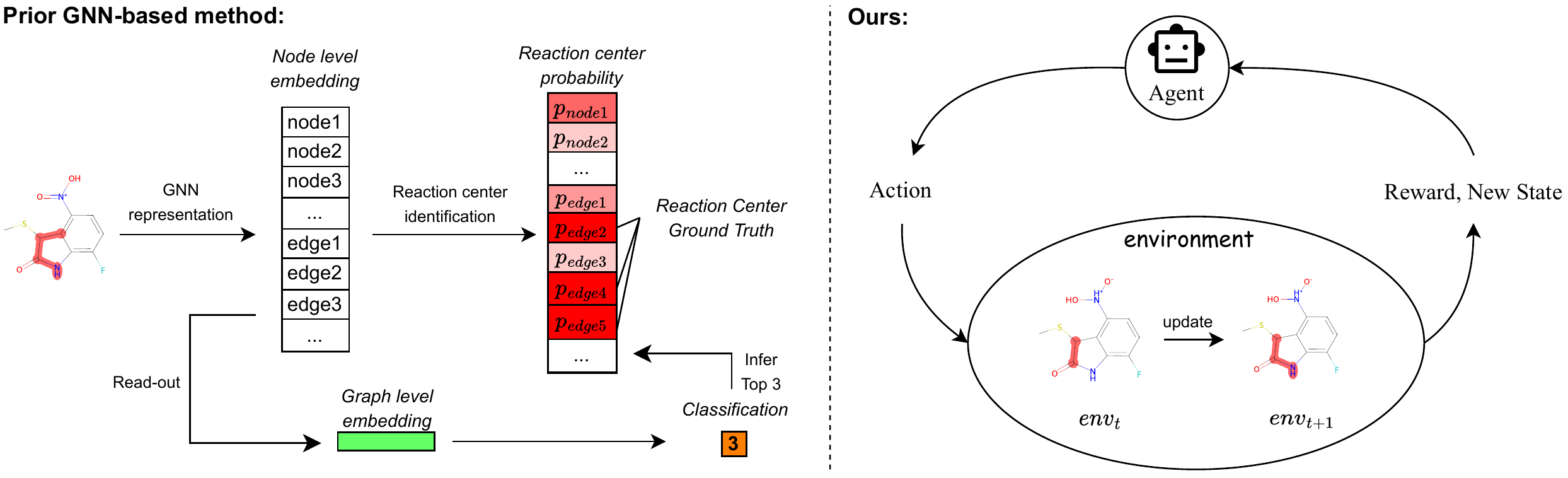}
	\caption{\textbf{Left:} An illustration of Prior GNN-based method. \textbf{Right:} An overview of our proposed RCsearcher.}
	\label{intro2}
\end{figure*}

Prior studies on reaction center identification mainly appear in semi-templated retrosynthesis methods. Moreover, they are limited to single reaction center identification. Here, a single reaction center refers to the reaction center composed of a bond or an atom (Fig. \ref{intro1}(a)). G2G  \cite{shi2020graph} and RetroXpert  \cite{yan2020retroxpert} regards the bond that appears in the product but not in the reactants as the reaction center. They use R-GCN \cite{schlichtkrull2018modeling} and EGAT \cite{kaminski2022rossmann} respectively to predict the probability of each bond as a single reaction center. GRAPHRETRO \cite{somnath2021learning} also considers the atom whose number of attached hydrogens changes in the product as a single reaction center. Then, it uses MPN \cite{gilmer2017neural} to model the probability of each bond or atom as a single reaction center. Despite the excellent performance of semi-templated retrosynthesis methods, they are still not widely used. The main reason is that the current reaction center identification method cannot detect multiple reaction center effectively.

In reality, many reaction centers in reactions' products are composed of multiple bonds or atoms, which is not fully represented in the USPTO-50k dataset. In this paper, we refer to it as the multiple reaction center (Fig. \ref{intro1}(a)). \cite{dai2019retrosynthesis} creates a large dataset USPTO-full from the entire set of USPTO 1976-2016 (1,808,937 raw reactions). USPTO-full has roughly 1M unique reactions, and the proportion of samples with multiple reaction center reaches 38.84\%. For example, the reaction center of the product of most ring-opening reactions is the multiple reaction center \cite{coley2019rdchiral}. In retrosynthesis, the prior models sometimes misidentify multiple reaction center as the single reaction center leading to predictions deviating from the ground truth. It results in the same intermediate molecules being visited many times in the process of multi-step retrosynthesis \cite{xie2022retrograph}. As shown in Fig. \ref{intro1}(b), due to misidentifying multiple reaction center as single reaction center, the ring-opening operation is not performed, which leads to an infinite loop in multi-step retrosynthesis.

To the best of our knowledge, the only existing method that can simultaneously address single and multiple reaction center identification task appear in RetroXpert \cite{yan2020retroxpert}. It uses one shared EGAT \cite{kaminski2022rossmann} to output two tensors (Fig. \ref{intro2}). One tensor represents the probability that each bond is the reaction center. Another tensor indicates the result of multi-classification, where each category indicates the number of bonds forming the reaction center. During inference, the prediction of the multi-classification task is used to select the bond corresponding to the highest probability as the reaction center. There is a gap between its two joint optimization objectives and the purpose of finding the right single or multiple reaction center. Hence, designing a unified and reasonable framework for single and multiple reaction center identification remains a challenge.

In this paper, we present RCsearcher, a unified framework for single and multiple reaction center identification that combines the advantages of graph neural network and deep reinforcement learning (Fig. \ref{intro2}). The critical insight in this framework is that the single or multiple reaction center must be a node-induced subgraph of the molecular product graph \cite{coley2019rdchiral}. RCsearcher represents states and actions in continuous embeddings by graph representation learning \cite{kaminski2022rossmann} and uses a Deep Q-Network (DQN) \cite{mnih2013playing} to predict action distributions. At each step, it considers choosing one node in the molecular product graph and adding it to the explored node-induced subgraph as an action. After satisfying the stopping condition, RCsearcher regards the explored node-induced subgraph as the predicted reaction center. For effective and fair evaluation, we uniformly sample 40k reactions from USPTO-full to build a more general dataset USPTO-40k, where the proportion of samples with multiple reaction centers is also 38.84\%. Comprehensive experiments demonstrate that RCsearcher consistently outperforms other baselines for the reaction centre identification task, and it can extrapolate the reaction center patterns that have not appeared in the training set. Ablation experiments verify the effectiveness of individual components, including the beam search and one-hop-constraint of action space. In brief, we highlight our main contributions as follows:

\begin{itemize}
\item We address the important yet challenging task of reaction center identification and propose a unified framework RCsearcher both for single and multiple reaction center as the solution.
\item The key novelty is the GNN-based DQN which provides a strategy to select the node-induced subgraph as the predicted reaction center, and the core insight in this framework is that single or multiple reaction center must be a node-induced subgraph of the molecular product graph.
\item For fair and effective evaluation, we build a more general dataset USPTO-40k. We conduct extensive experiments to demonstrate the effectiveness and generalization of the RCsearcher. Ablation experiments also verify the effectiveness of individual components.
\end{itemize}

\section{Problem Definition}
The reaction center (RC) consists of atoms in the product whose local properties are not identical to the corresponding atoms in the reactants \cite{coley2019rdchiral}. In this paper, we use RDChiral \cite{coley2019rdchiral} to extract the super general reaction center. Hence, regardless of whether the single or multiple reaction center must be a node-induced subgraph of the molecular product graph.

\textbf{Notation} A molecular product graph $\mathcal{G}_{p}=(\mathcal{V}_{p}, \mathcal{E}_{p})$ is represented as a set of $|\mathcal{V}_{p}|$ nodes (atoms) and a set of $|\mathcal{E}_{p}|$ edges (bonds). The reaction center graph $\mathcal{G}_{rc}=(\mathcal{V}_{rc}, \mathcal{E}_{rc})$ is a node-induced subgraph of the molecular product graph $\mathcal{G}_{p}$ such that $\mathcal{V}_{rc} \subseteq \mathcal{V}_{p}$ and $\mathcal{E}_{rc}=\left\{(u, v) \mid u, v \in \mathcal{V}_{rc},(u, v) \in \mathcal{E}_{p}\right\}$.

\textbf{Reaction Center Identification} Given the molecular product graph $\mathcal{G}_{p}$, reaction center identification aims to detect the corresponding reaction center graph $\mathcal{G}_{rc}$, i.e. $\mathcal{G}_{rc}$'s node set $\mathcal{V}_{rc}$.

\textbf{Evaluation} There may be several isomorphic node-induced subgraphs in a graph. Therefore, the condition for correct identification is that the node set of the predicted reaction center graph is consistent with the ground-truth $\mathcal{V}_{rc}$.

\section{Related work}
\textbf{Efforts on Reaction Center Identification in Retrosynthesis} \quad G2G  \cite{shi2020graph} and RetroXpert  \cite{yan2020retroxpert} regards the bond that appears in the product but not in the reactants as the reaction center. G2G uses R-GCN \cite{schlichtkrull2018modeling} to predict the probability of each bond as a single reaction center. In contrast, RetroXpert uses EGAT \cite{kaminski2022rossmann} to model the probability of each bond as the reaction center, and it adds a graph-level auxiliary task to predict the total number of disconnection bonds (Fig. \ref{intro2}). Hence, RetroXpert is not limited to single reaction center identification. GRAPHRETRO \cite{somnath2021learning} also considers the atom whose number of attached hydrogens changes in the product as a single reaction center. Then, it uses MPN \cite{gilmer2017neural} to model the probability of each bond or atom as a single reaction center. It is worth noting that GRAPHRETRO roughly proposed an autoregressive model for multiple reaction center identification in its appendix. However, GRAPHRETRO does not elaborate on the specific details of the autoregressive model, and the code of the autoregressive model is not available in the official GitHub link of GRAPHRETRO (https://github.com/vsomnath/graphretro).

\textbf{Efforts on Single-step Retrosynthesis Prediction} \quad Single-step prediction models can be categorized into three main classes, i.e., template-based, template-free and semi-template-based method. Template-based methods rely on templates that encode the core of chemical reactions, converting product molecules into reactants. The key is to rank the templates and select the appropriate one to apply, and recent attempts \cite{coley2017computer, dai2019retrosynthesis} have addressed the template selection problem by similarity and neural networks. Despite their superior interpretability, template-based approaches are disadvantaged by poor generalization to structures beyond templates. On the other hand, template-free methods \cite{liu2017retrosynthetic, schwaller2019molecular} regard single-step retrosynthesis prediction as a translation task and translate a product molecule represented in SMILES strings \cite{weininger1988smiles} to reactant SMILES strings. To combine the benefits of template-based and template-free methods, recent works \cite{shi2020graph, yan2020retroxpert, somnath2021learning} seek semi-template-based methods where they first predict single reaction center via graph neural networks and the then intermediate synthons are secondly translated into reactants via seq2seq or graph translation models.

\textbf{Efforts on Multi-step Retrosynthetic planning} \quad HgSearch \cite{schwaller2020predicting} and Proof Number Search \cite{kishimoto2019depth} are traditional heuristic search algorithms in which chemical feasibility and failed pathway values are not considered. \cite{segler2018planning} adopt Monte Carlo tree search to generate search trees and explore multiple synthetic paths dynamically. \cite{chen2020retro} devise a neural-based A*-like algorithm that learns an additional value network with automatically constructed and only successful paths to bias the search prior. \cite{han2022gnn} and \cite{xie2022retrograph} use a GNN-based value network to capture intermolecular/intra-pathway level information to further improve A*-like retrosynthesis planning algorithms. Notably, \cite{xie2022retrograph} observe that the same intermediate molecules are visited many times in the searching process. Sometimes, the ring-opening operation is not performed due to misidentifying multiple reaction center as single reaction center, which leads to an infinite loop in multi-step retrosynthesis (Fig. \ref{intro1}(b)).

\section{Preliminaries and Proposed Method}\label{s4}
In this section, we present our RL based reaction center identification in retrosynthesis method, RCsearcher. The rest of Section \ref{s4} is organized as follows. Section \ref{Preliminaries} introduces the preliminaries. Section \ref{Overview} presents a high-level overview of how to leverage Deep Q-Network (DQN) \cite{mnih2013playing} to tackle the reaction center identification in retrosynthesis (Fig. \ref{model}). Section \ref{Details} describes the details of the state, action, reward (Fig. \ref{model} \textbf{Left}). Section \ref{Training} explains how to train the DQN efficiently. Section \ref{Inference} shows how RCsearcher does inference with beam search mechanism (Fig. \ref{model} \textbf{Right}).  Section \ref{complexity} analyzes the time complexity.

\subsection{Preliminaries}\label{Preliminaries}
In this paper, we adopt the EGAT \cite{kaminski2022rossmann} to compute both the node embeddings and the edge embedding. Given $\mathbf{H}^{(t)}= [ \boldsymbol{h}_{1}^{(t)}; \boldsymbol{h}_{2}^{(t)}; \cdots ; \boldsymbol{h}_{n}^{(t)}] \in R^{n \times d}$ and $\mathbf{f}_{i j}^{(t)}\in R^{1 \times d}$ are the node embedding matrix and the embedding of edge $(i, j)$ at the $t$-th layer repectively, where $\boldsymbol{h}_{i}^{(t)}\in R^{1 \times d}$ is the node-level embedding for node $i$ of the graph and is also the $i$-th row of $\mathbf{H}^{(t)}$, $d$ is the dimension of node-level embedding and $n$ is the number of nodes. $\boldsymbol{h}_{i}^{(0)}$ and $\boldsymbol{f}_{ij}^{(0)}$ are the initial features of node and edge in molecular product graph $\mathcal{G}_{p}$. The details of the initial features can be found in the appendix. EGAT injects  the graph structure into the attention mechanism by performing masked attention, namely it only computes $\alpha_{i j}$ for nodes $j \in \mathcal{N}_{i}$, where $\mathcal{N}_{i}$ is the first-order neighbors of node $i$ in the graph:
\begin{equation}
	\label{egat}
	\begin{aligned}
        \boldsymbol{f}_{i j}^{(t+1)}&=\text { LeakyReLU }\left(\left[\boldsymbol{h}_i^{(t)}\mathbf{W}\left\|\boldsymbol{f}_{i j}^{(t)}\right\| \boldsymbol{h}_j^{(t)}\mathbf{W}\right]A\right), \\
		e_{ij} &= \mathbf{a} \cdot {\boldsymbol{f}_{i j}^{(t+1)}}^{T}, \alpha_{i j} =\frac{\exp \left(e_{ij}\right)}{\sum_{k \in \mathcal{N}_{i}} \exp \left(e_{ik}\right)},
	\end{aligned}
\end{equation}
where $e_{ij}\in R$ and $\alpha_{i j}\in R$ are a non-normalized attention coefficient and a normalized attention coefficient representing the weight of message aggregated from node $j$ to node $i$ respectively in the $t$-th layer of EGAT, and $\|$ is the concatenation operation. Besides,$\mathbf{W}\in R^{d \times d}$, $A\in R^{3d \times d}$ and $\mathbf{a}\in R^{1 \times d}$ are learnable parameters in the $t$-th layer.

EGAT employs multi-head attention to stabilize the learning process of self-attention, similar to Transformer \cite{vaswani2017attention}. If there are $K$ heads, $K$ independent attention mechanisms execute the Eq. \ref{egat}, and then their features are concatenated:
\begin{equation}
	\label{att}
	\begin{aligned}
	    \boldsymbol{h}_{i}^{(t+1)} &= \operatorname{MLP} \left(\|_{k=1}^{K} \sigma\left(\sum_{j \in \mathcal{N}_{i}} \alpha_{i j}^{k}  \boldsymbol{h}_{j}^{(t)} \mathbf{W}^{k}\right)\right)
	\end{aligned}
\end{equation}
where $\|$ represents concatenation, $\alpha_{i j}^{k}$ are normalized attention coefficients computed by the $k$-th learnable $\mathbf{W}^{k}\in R^{d \times d}$, $A^{k}\in R^{3d \times d}$ and $\mathbf{a}^{k}\in R^{1 \times d}$ following Eq. \ref{egat}. Besides, $\operatorname{MLP}$ denotes multi-perceptron.  For simplicity, we denote the encoding process by $\mathrm{EGAT}(\cdot)$ in this study.

\subsection{Overview of RCSeacher}\label{Overview}
RCSeacher enables graph representation learning techniques to tackle the reaction center identification in retrosynthesis and uses deep Q-learning to select one node that is added to the current explored subgraph in each search state. RCSeacher represents states and actions in continuous embeddings, and maps $(s_{t}, a_{t})$ to a score $Q(s_{t}, a_{t})$ via a DQN which consists of a Graph Neural Network encoder and learnable components to project the
representations into the final score. Here, $s_{t}$ and $a_{t}$ denote the state and action respectively in the step $t$.  Once RCSeacher is trained, it can be applied to any new graphs that are unseen during training.

\begin{figure*}[t]
	\centering
 \label{model}
	\includegraphics[width=1\linewidth]{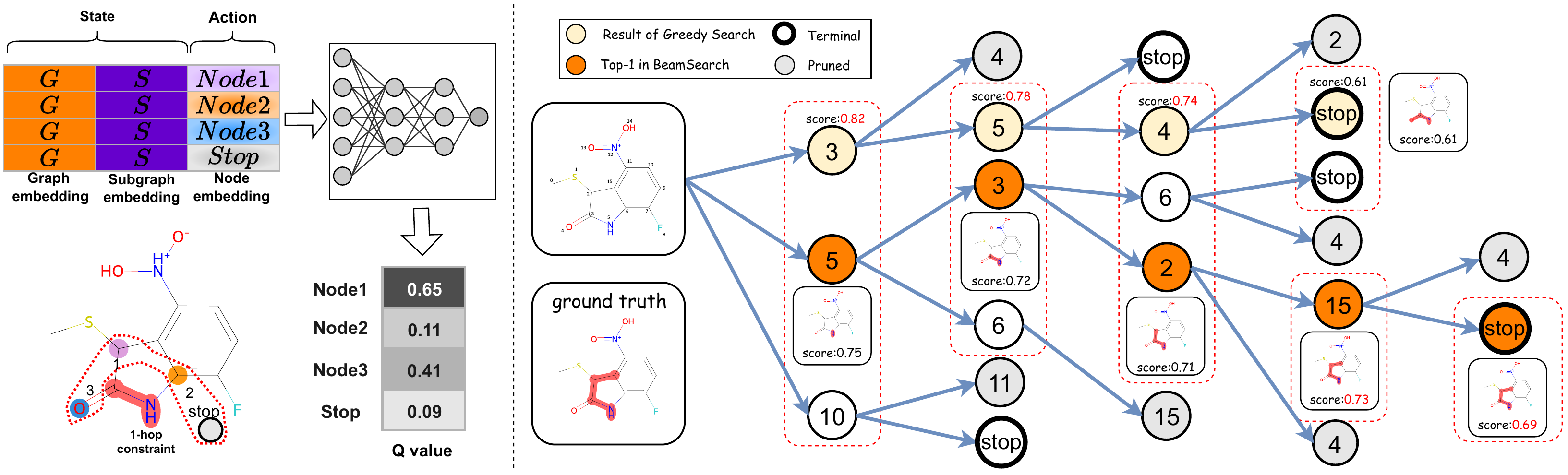}
	\caption{\textbf{Left:} The details of Q-NET, and representation of state and action. The dashed red line indicates the one-hop constraint. \textbf{Right:}  An illustration of Beam Search during inference.}
	\label{intro2}
\end{figure*}

State $s_{t}$ consists of the (1)
the molecular product graph $\mathcal{G}_{p}$, (2) current explored reaction center graph $\hat{\mathcal{G}}_{rc}^{t} = \mathcal{G}_{p}[\hat{\mathcal{V}}_{rc}^{t}]$. Action $a_{t}$ is defined as selecting one node from the first-order neighbor of the current explored subgraph $\hat{\mathcal{V}}_{rc}^{t}$, or a stop action implying stop of the exploration process. Since most reaction center graphs have only one connected branch, we impose a one-hop constraint on the action space to narrow the search space and improve performance. For RCSeacher, given our goal, the intermediate reward is defined as $r_{t} = 0$ at any intermediate step $t$ and $r_{T} = 1$ when predicted node-set $\hat{\mathcal{V }}_{rc}^{T}$ is consistent with ground-truth $\mathcal{V}_{rc}$, otherwise $r_{T} = 0$ at the final step $T$. Therefore, our optimization goal, maximizing the expected remaining future reward, means finding a reaction center that is exactly the same as the ground-truth.

Since the action space can be large, we leverage the representation learning capacity of continuous representations for DQN design. At state $s_{t}$, for each action $a_{t}$, our DQN predicts a $Q(s_{t}, a_{t})$ representing the remaining future reward after selecting action $a_{t}$. Based on the above insights, we can design a simple
DQN leveraging the graph encoder EGAT \cite{kaminski2022rossmann} to obtain one embedding per node, $\left\{\boldsymbol{h}_i \mid \forall i \in \mathcal{V}_{p}\right\}$. Denote $\|$ as concatenation, \text{READOUT} as a readout operation that aggregates node-level embeddings into subgraph embeddings $\boldsymbol{h}_{\hat{\mathcal{G}}_{rc}^{t}}$, and whole-graph embedding $\boldsymbol{h}_{\mathcal{G}_{p}}$. A state can then be represented as $\boldsymbol{h}_{s_{t}} = \boldsymbol{h}_{\mathcal{G}_{p}} \| \boldsymbol{h}_{\hat{\mathcal{G}}_{rc}^{t}}$. An action can be represented as $\boldsymbol{h}_{a_{t}} \in \left\{\boldsymbol{h}_i \mid \forall i \in \mathcal{V}_{p}\right\} \cup \{\boldsymbol{h}_{stop} \}$, where $\boldsymbol{h}_{stop}$ denotes the stop action. The $Q$ function would then be designed as:
\begin{equation}
\label{q_function}
\begin{aligned}
&\,\,\,\,\,\,\,\,Q\left(s_t, a_t\right) \\ &=\operatorname{MLP}\left(\boldsymbol{h}_{s_t} \| \boldsymbol{h}_{a_t}\right) \\
&=\operatorname{MLP}\left(\boldsymbol{h}_{\mathcal{G}_{p}} \| \boldsymbol{h}_{\hat{\mathcal{G}}_{rc}^{t}} \| \boldsymbol{h}_{i}\right) or \operatorname{MLP}\left(\boldsymbol{h}_{\mathcal{G}_{p}} \| \boldsymbol{h}_{\hat{\mathcal{G}}_{rc}^{t}} \| \boldsymbol{h}_{stop}\right),
\end{aligned}
\end{equation}
where $\operatorname{MLP}$ denotes multi-perceptron.

\subsection{Details of State, Action and Reward}\label{Details}
\textbf{State} State $s_{t}$ consists of the (1) the molecular product graph $\mathcal{G}_{p}$, (2) current explored reaction center graph $\hat{\mathcal{G}}_{rc}^{t} = \mathcal{G}_{p}[\hat{\mathcal{V}}_{rc}^{t}]$. Firstly, we use EGAT \cite{kaminski2022rossmann} to obtain the graph-level embedding $\boldsymbol{h}_{\mathcal{G}_{p}} \in R^{1 \times d}$ representing the molecular product graph $\mathcal{G}_{p}$:
\begin{equation}
\label{product1}
\begin{aligned}
\left\{\boldsymbol{h}_i \mid \forall i \in \mathcal{V}_{p}\right\}, \left\{\boldsymbol{f}_{ij} \mid \forall (i,j) \in \mathcal{E}_{p}\right\} = \mathrm{EGAT}\left(\mathcal{G}_{p}\right),
\end{aligned}
\end{equation}
\begin{equation}
\label{product2}
\begin{aligned}
\boldsymbol{h}_{nodes} &= \operatorname{MEAN}\left(\left\{\boldsymbol{h}_i \mid \forall i \in \mathcal{V}_{p}\right\}\right), \\
\boldsymbol{h}_{edges} &= \operatorname{MEAN}\left(\left\{\boldsymbol{f}_{ij} \mid \forall (i,j) \in \mathcal{E}_{p}\right\}\right),
\end{aligned}
\end{equation}
\begin{equation}
\label{product3}
\begin{aligned}
\boldsymbol{h}_{\mathcal{G}_{p}}^{\prime}&=\operatorname{ABS}\left(\boldsymbol{h}_{nodes},  \boldsymbol{h}_{edges} \right) \| \left(\boldsymbol{h}_{nodes} + \boldsymbol{h}_{edges} \right), \\
\boldsymbol{h}_{\mathcal{G}_{p}} &= \operatorname{MLP}\left(\boldsymbol{h}_{\mathcal{G}_{p}}^{\prime}\right),
\end{aligned}
\end{equation}
where $\boldsymbol{h}_{nodes} \in R^{1 \times d}$ and $\boldsymbol{h}_{edges} \in R^{1 \times d}$ are the whole graph information from node and edge perspectives respectively. $\operatorname{ABS}$ denotes absolute difference and $\|$ refers to concatenation. It ensures our representations are permutation invariant. Besides, $\operatorname{MLP}$ denotes multi-perceptron.

Secondly, we use the above node-level embeddings $\left\{\boldsymbol{h}_i \mid \forall i \in \mathcal{V}_{p}\right\}$ to derive the subgraph embedding $\boldsymbol{h}_{\hat{\mathcal{G}}_{rc}^{t}} \in R^{1 \times d}$ representing the current explored reaction center graph $\hat{\mathcal{G}}_{rc}^{t}$:
\begin{equation}
\boldsymbol{h}_{\hat{\mathcal{G}}_{rc}^{t}}=
\left\{
\begin{aligned}
%\nonumber
&\vec{0}\,\,\,\,\,\,\,\,\,\,\,\,\,\,\,\,\,\,\,\,\,\,\,\,\,\,\,\,\,\,\,\,\,\,\,\,\,\,\,\,\,\,\,\,\,\,\,\,\,\,\,\,\,\,\,\,\,\,\,\,\,\,\,\,\,\,\,\,, t=0\\
&\operatorname{MEAN}\left(\left\{\boldsymbol{h}_i \mid \forall i \in \hat{\mathcal{V}}_{rc}^{t}\right\}\right)\,\,, t \not= 0\\
\end{aligned}
\right.
.
\end{equation}
Since the exploration process did not start at step $t=0$ ($\hat{\mathcal{V}}_{rc}^{0}=\emptyset$), we use $\vec{0} \in R^{ 1 \times d}$ to represent $\hat{\mathcal{V}}_{rc}^{0}$.

\textbf{Action} Since most reaction center graphs have only one connected branch, we impose a one-hop constraint on the action space to narrow the search space and improve performance. In other words, the action $a_t$ is to select one node from the first-order neighbour of the current explored subgraph $\hat{\mathcal{V}}_{rc}^{t}$ or a stop action implying stop of the exploration process. We use the node $i$'s embedding $\boldsymbol{h}_{i} \in R^{1 \times d}$ and one learnable embedding $\boldsymbol{h}_{stop} \in R^{1 \times d}$ to represent the action of the selecting one node and the stop action respectively:
\begin{equation}
\boldsymbol{h}_{a_t}=
\left\{
\begin{aligned}
%\nonumber
&\boldsymbol{h}_{i}\,\,\,\,\,\,\,\,\,\,\,\,\,\,\,\,\,\,\,\,\,\,\,, i \in \mathcal{V}_{p}, t=0\\
&\boldsymbol{h}_{i} \, or \, \boldsymbol{h}_{stop} \,\,, i \in \operatorname{ONE-HOP}\left(\hat{\mathcal{V}}_{rc}^{t}\right), t \not= 0\\
\end{aligned}
\right.
,
\end{equation}
where $\operatorname{ONE-HOP}$ is the operation of 1-hop constrain that can obtain the first-order neighbour of the current explored subgraph.

\textbf{Reward} When the action $a_t$ is the stop action, we refer to this step $t$ as the final step $T$. We define the reward $r_t$ as:
\begin{equation}
r_{t}=
\left\{
\begin{aligned}
%\nonumber
&0\,\,\,\,\,\,\,\,\,\,\,\,\,\,\,\,\,\,\,\,\,\,\,,t=0,1,2, \cdots, T-1\\
&0\,\,\,\,\,\,\,\,\,\,\,\,\,\,\,\,\,\,\,\,\,\,\,, \hat{\mathcal{V}}_{rc}^{T} \not= \mathcal{V}_{rc},t=T\\
&1\,\,\,\,\,\,\,\,\,\,\,\,\,\,\,\,\,\,\,\,\,\,\,, \hat{\mathcal{V}}_{rc}^{T} = \mathcal{V}_{rc},t=T\\
\end{aligned}
\right.
.
\end{equation}
Therefore, our optimization goal, maximizing the expected remaining future reward, means finding a reaction center that is exactly the same as the ground truth.

\subsection{Training}\label{Training}
We adopt the standard Deep Q-learning framework \cite{mnih2013playing}. Since imitation learning is known to help with training stability and performance \cite{levine2013guided}, we allow the agent to follow ground-truth trajectories. For each experience $(s_t,a_t,r_t,s_{t+1})$ in the experience pool, our loss function is:
\begin{equation}
	\begin{aligned}
	 loss &= (y_t - Q(s_t, a_t))^2,\\
  y_t &= \gamma r_t + \operatorname{MAX}\left(Q(s_{t+1}, a_{t+1})\right),
	\end{aligned}
\end{equation}
where $\gamma \in (0,1]$ is the decay rate, and $y_t$ is a constant and \textbf{NO-GRAD}.

\subsection{Inference}\label{Inference}
We use the beam search mechanism \cite{meister2020if} to derive the $k$ best predictions for inference. With a hyperparameter budget BEAM SIZE $k$, the agent can transition to at most BEAM SIZE $k$ number of best new states at any given state. Thus, we have $k^2$ candidate states before the next level in the beam search process. From the $k^2$ candidate states, we select the $k$ best states according to the predicted value of the $Q$ function at the next level. For example, in Fig. \ref{model}, BEAM SIZE = 3, each level of the beam search can have up to 3 state nodes (red dashed box). The pseudocode is shown in Alg. \ref{beamsearch}.

\begin{algorithm}
    \caption{Inference with Beam Search}
    \label{beamsearch}
\begin{algorithmic}
    \STATE {\bfseries Input:} $Q$ function $Q(s, a)$; Initial State $s_0$; Beam Size $k$
    \STATE {\bfseries Output:} $k$ best results $\textsc{K-results}$
    \STATE $\textsc{beam} = \{s_0\}$
    \STATE $\textsc{candidate} = \emptyset$
    \REPEAT
    \STATE $\textsc{K-results} = \emptyset$
        \FOR{$s$ {\bfseries in} \textsc{beam}}
            \STATE $\textsc{Top-K-actions} = \underset{a}{\operatorname{argtopK}}\left(Q(s, a)\right)$
            \FOR{$a$ {\bfseries in} $\textsc{Top-K-actions}$}
            \STATE add $(s, a)$ to $\textsc{candidate}$
            \ENDFOR   
        \ENDFOR
        \STATE $\textsc{beam}^{\prime} = \underset{(s, a) \in \textsc{candidate}}{\operatorname{argtopK}}\left(Q(s, a)\right)$
        \STATE $\textsc{candidate} = \emptyset$
        \STATE $\textsc{beam} = \emptyset$

        \FOR{$(s, a)$ {\bfseries in} $\textsc{beam}^{\prime}$}
        \STATE $s^{\prime} \leftarrow $ execute $a$ on $s$
        \IF{$a$ is the stop action}
        \STATE add $(s, a)$ to $\textsc{candidate}$
        \STATE add $s$ to $\textsc{K-results}$
        \ELSE
        \STATE add $s^{\prime}$ to $\textsc{beam}$
        \ENDIF
        \ENDFOR
    \UNTIL{$\textsc{beam} = \emptyset$}
    \STATE \textbf{Return:} $\textsc{K-results}$
\end{algorithmic}
\end{algorithm}

\subsection{Complexity Analysis}\label{complexity}
At each state, the agent calculates the future values which have worst-case time complexity $\mathcal{O}\left(|\mathcal{V}_{p}|\right)$ due to the action space consisting of all atoms in the product at step 0. Overall the beam search depth is bounded by $|\mathcal{V}_{p}|$, and each level of the beam search has at most BEAM SIZE $k$ states. Thus, the overall time complexity is $\mathcal{O}\left( 
 k \times  |\mathcal{V}_{p}|^{2}   \right)$.

\section{Evaluation}
In this section, we evaluate the performance of our RCsearcher with a comparison to some baseline approaches for the reaction center identification in retrosynthesis, and with significant goals of addressing the following questions: {\bf Q1: }How effective and robust is RCsearcher compared to the baseline approaches? {\bf Q2: }How well does RCsearcher perform on the data with the multiple reaction center? {\bf Q3: }How does the proposed one-hop constrain and beam search in inference improve performance? {\bf Q4: }Does RCsearcher have more robust generalization than baselines?

\begin{figure}[t]
	\centering
	\includegraphics[width=1\linewidth]{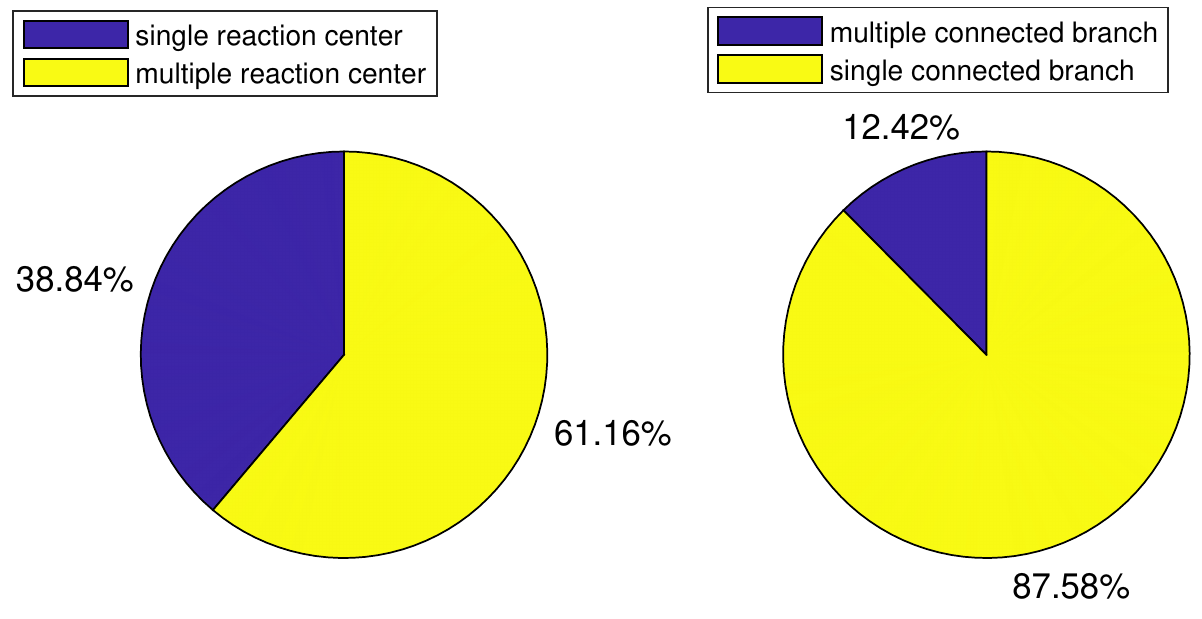}
	\caption{The distribution of the USPTO-40k.}
	\label{pie}
\end{figure}

\textbf{Data} Since the data with the multiple reaction center in USPTO-50k only accounts for about 5\%, for effective evaluation, we created a new dataset USPTO-40k. We adopt stratified random sampling for USPTO-full and take out 40k samples, ensuring that the data distribution is close enough to USPTO-full. In the USPTO-40k, the data with the multiple reaction center accounts for 38.84\%. In the USPTO-40k, 12.42\% of the samples' reaction center consists of more than one connected branch. We randomly select 80\% of the samples as the training set and divide the rest into validation and test sets with equal sizes. Fig. \ref{pie} shows the distribution of the dataset. The details of the USPTO-40k can be found in the appendix. 

\begin{table*}[t]
\centering
\caption{Top-k accuracy for Reaction Center Identification on USPTO-40K. The notations '-20' and '-30' denote 20 sampling experiments and 30 sampling experiments respectively.}
\label{overall}
\begin{tabular}{ll|rrrr}
\hline
\multicolumn{2}{c|}{Methods} &
  \multicolumn{1}{c}{Top-1(\%)} &
  \multicolumn{1}{c}{Top-2(\%)} &
  \multicolumn{1}{c}{Top-3(\%)} &
  \multicolumn{1}{c}{Top-4(\%)} \\ \hline
\multicolumn{1}{l|}{\multirow{3}{*}{GNN-based}}        & MPN-based         & 9.30$\pm$0.56  & 10.28$\pm$0.38 & 11.58$\pm$0.31 & 14.07$\pm$0.14 \\
\multicolumn{1}{l|}{}                                  & EGAT-based        & 18.04$\pm$0.79 & 23.49$\pm$0.66 & 25.33$\pm$0.43 & 25.78$\pm$0.25 \\
\multicolumn{1}{l|}{}                                  & RGCN-based        & 9.27$\pm$0.46  & 14.25$\pm$0.47 & 17.53$\pm$0.35 & 20.78$\pm$0.17 \\ \hline
\multicolumn{1}{l|}{\multirow{4}{*}{Seq2Seq}} &
  Product2RC-20 & 3.25$\pm$0.51 & 4.95$\pm$0.34 & 6.45$\pm$0.28 & 8.68$\pm$0.19 \\
\multicolumn{1}{l|}{} &
  Product2RC-30 & 3.15$\pm$0.42 & 4.61$\pm$0.33 & 6.42$\pm$0.20 & 8.48$\pm$0.13 \\
\multicolumn{1}{l|}{}                                  & Product2RC-Aug-20 & 7.78$\pm$0.37  & 10.22$\pm$0.20 & 12.70$\pm$0.16 & 15.00$\pm$0.06 \\
\multicolumn{1}{l|}{}                                  & Product2RC-Aug-30 & 8.51$\pm$0.40  & 10.36$\pm$0.46 & 13.33$\pm$0.29 & 15.85$\pm$0.22 \\ \hline
\multicolumn{1}{l|}{\multirow{2}{*}{Similarity-based}} & Sim-based-20      & 19.51$\pm$0.41 & 23.33$\pm$0.20 & 26.54$\pm$0.17 & 27.04$\pm$0.06 \\
\multicolumn{1}{l|}{}                                  & Sim-based-30      & 19.58$\pm$0.39 & 23.21$\pm$0.20 & 26.81$\pm$0.19 & 27.13$\pm$0.07 \\ \hline
\multicolumn{1}{l|}{Ours}                              & RCsearcher          & \textbf{30.75$\pm$0.41} & \textbf{31.45$\pm$0.35} & \textbf{31.92$\pm$0.28} & \textbf{32.38$\pm$0.18} \\ \hline
\end{tabular}
\end{table*}

\textbf{Baselines} We compare RCsearcher to three baselines for evaluating overall performance: GNN-based, Seq2Seq, and Similarity-based methods. These include:

(1) GNN-based: We use the method in RetroXpert. It uses one graph encoder to model the probability of each bond as the reaction center, and adds a graph-level auxiliary task to predict the total number of disconnection bonds. Here, we use three graph encoder, including R-GCN \cite{schlichtkrull2018modeling}, EGAT \cite{kaminski2022rossmann}, MPN \cite{gilmer2017neural}. Thus, we refer to these three baselines as \textbf{RGCN-based}, \textbf{EGAT-based} and \textbf{MPN-based}.

(2) Seq2Seq: Here, the SMILES string of the product is the input sequence, and the SMARTS string of the reaction center is the input sequence. We use Transformer \cite{vaswani2017attention} as the neural sequence-to-sequence model. We follow the same data augmentation tricks used in \cite{wan2022retroformer} for the SMILES generative models. Instead of
expanding the training dataset off-the-shelf, we perform the augmentation on-the-fly. At each iteration, there is a probability of 50\% to permute the SMILES. We refer to this baseline as \textbf{Product2RC} and \textbf{Product2RC-Aug}.

(3) Similarity-based: We calculate the similarity between the input product and each product in the training set and then sort the products in training set based on the similarity. We regard the reaction center of the corresponding product in training set as the prediction. Here, we use molecular morgan fingerprints to calculate similarity. We refer to this baseline as \textbf{Sim-based}.

It is worth noting that \textbf{Product2RC} and \textbf{Sim-based} can only predict the reaction center pattern and cannot accurately obtain the specific position of the reaction center in the product. We regard the reaction center pattern as the query graph and perform subgraph matching on the product. When there is only one solution in the matching result, the only solution is regarded as the predicted result, otherwise, we randomly select a solution as the predicted result. In this way, we can calculate the top-k results of the overall test set once. We repeat this experiment N times, and we finally take the average top-k accuracy of these N times as the final result.

\textbf{Evaluation Metric} For different methods, we transform the predicted results into the predicted reaction center graph's node set $\hat{\mathcal{V}}_{rc}^{T}$. The condition for correct identification is that the node set $\hat{\mathcal{V}}_{rc}^{T}$ of the predicted reaction center graph is consistent with the ground-truth $\mathcal{V}_{rc}$. We use the top-k exact match accuracy as our evaluation metrics.

\textbf{Implementation Settings} Our proposed RCsearcher is implemented with Deep Graph Library (DGL) \cite{wang2019dgl} and Pytorch \cite{paszke2019pytorch}. As for the EGAT, we stack four identical four-head attentive layers of which the hidden dimension is 256. All embedding sizes in the model are set to 256. We use $\operatorname{ELU}(x)=\alpha(\exp (x)-1)$ for $x \leq 0$ and $x$ for $x > 0$ as our activation function where $\alpha=1$. We conduct all the experiments on a machine with an Intel Xeon 4114 CPU and one Nvidia Titan GPU. For training, we set the learning rate to 0.001, the number of training iterations to 100000, and use the Adam optimizer \cite{kingma2014adam}. The first 10000 iterations are imitation learning. Checkpoints are saved for each 1000 iteration to select the best checkpoints on the evaluation set. The source code can be found in the supplementary materials.

\subsection{Overall Performance} 
We operate the training process five times and report the mean and standard deviation in accuracy. The overall performance is illustrated in Table \ref{overall}. We have two observations. {\bf First}, our method achieves state-of-the-art performance on the dataset. It indicates that our method can better handle single and multiple reaction center identification. {\bf Second}, the average performance of \textbf{Product2RC} and \textbf{Sim-based} with 20 and 30 sampling
 is close. It implies that the average of 30 results is enough to fairly reflect the capabilities of \textbf{Product2RC} and \textbf{Sim-based}.

\subsection{Performance on Multiple Reaction Center}
\begin{figure}[t]
	\centering
	\includegraphics[width=1\linewidth]{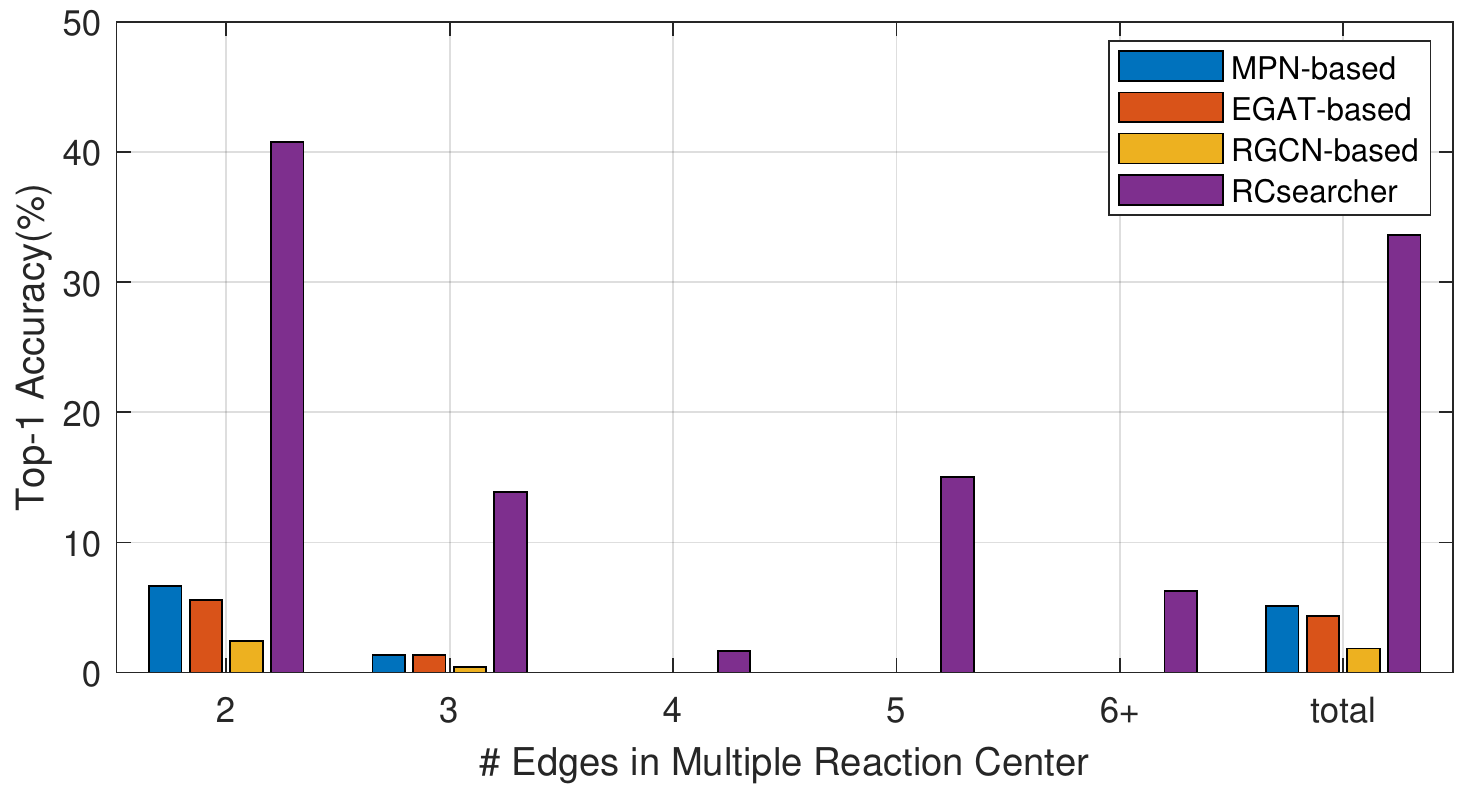}
	\caption{The performance of the multiple reaction center identification.}
	\label{multi}
\end{figure}

To address the {\bf Q2}, we train RCsearcher on the train data with multiple reaction center to explore its performance on the test data with multiple reaction center. The multiple reaction center identification results are shown in Fig. \ref{multi}. The overall top-1 accuracy of Rcsearcher is about eight times higher than that of prior GNN-based methods. Also, the prior GNN-based methods completely fail on data with the reaction center consisting of more than two edges. Conversely, RCsearcher still works. It shows that RCsearcher's performance on multiple reaction center identification is significantly better than baselines. We attribute it to consistency between our optimization objectives and the purpose of finding the right single or multiple reaction center.

\subsection{Performance of Generalization}
\begin{figure}[t]
	\centering
	\includegraphics[width=1\linewidth]{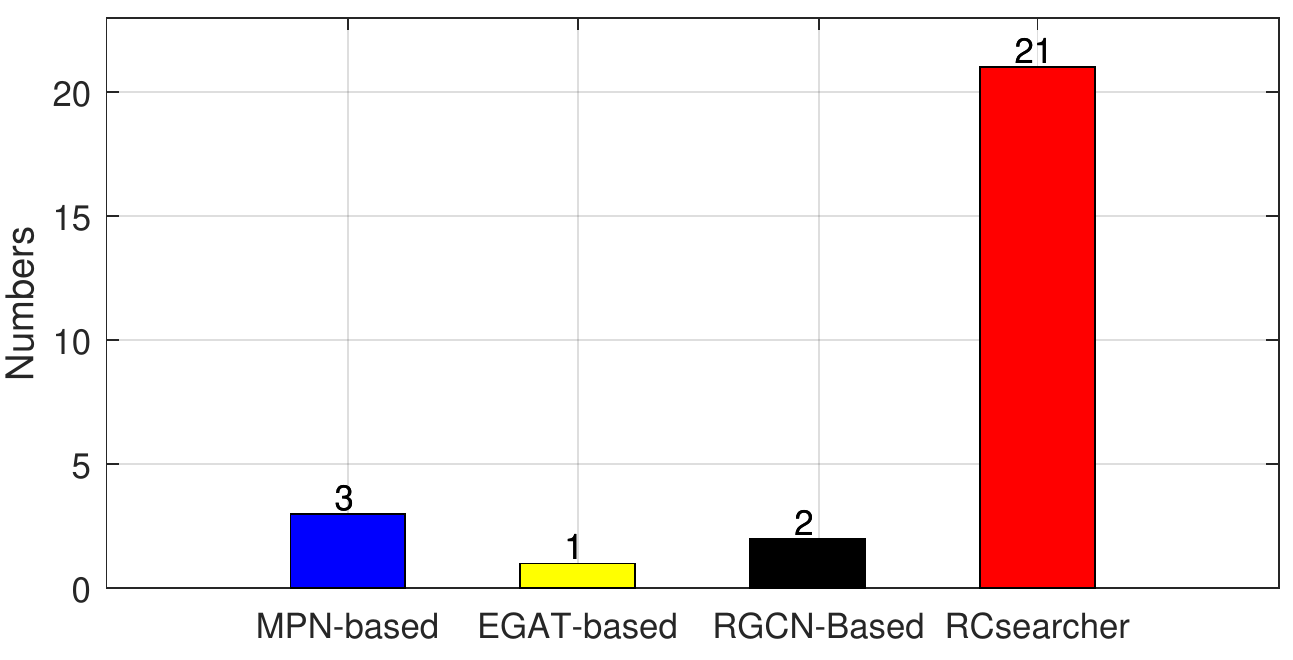}
	\caption{The performance of the extrapolations.}
	\label{generative}
\end{figure}

In order to evaluate the generalization ability of the RCsearcher, we count the number of reaction center patterns that are extrapolated successfully and not in the training set. The results are shown in Figure \ref{generative}. The number of extrapolations of Rcsearcher is about twenty times higher than that of prior GNN-based methods. It demonstrates that our method can predict novel reaction center patterns that have not appeared in the training set.

\subsection{Ablation Study}
The results of the ablation study are illustrated in Table \ref{ablation}, and on which we have the two observations: {\bf 1)} The setting with the beam search results is 0.52 percentage points higher than the setting without the beam search, which means beam search can improve performance. {\bf 2)} The setting with the one-hop constrain results are 3.62 percentage points higher than the setting without the one-hop constrain. It demonstrates that although the one-hop constraint fails in some samples with reaction center of multi-connected branches, it can improve the model's performance.

\begin{table}[h]
\centering
\caption{The Results of Ablation Study.}
\label{ablation}
\begin{tabular}{l|c}
\hline
\multicolumn{1}{c|}{Ablation Setting} & Top-1 Accuracy \\ \hline
w/ beam search                        & 30.75\%              \\
w/o beam search                       & 30.23\%              \\ \hline
w/ one-hop constrain                  & 30.75\%              \\
w/o one-hop constrain                 & 27.13\%              \\ \hline
\end{tabular}
\end{table}

\subsection{Hyperparameter Sensitivity Analysis}
\begin{figure}[t]
	\centering
	\includegraphics[width=1\linewidth]{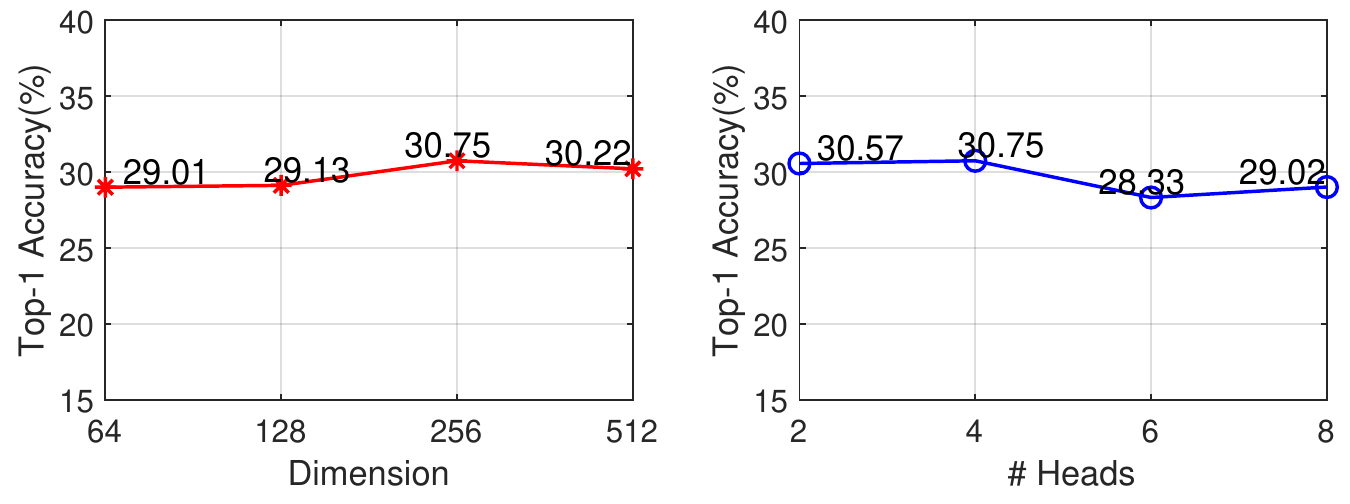}
	\caption{Hyperparameter Sensitivity Analysis.}
	\label{robustness}
\end{figure}

We respectively fix the number of heads to 4 and the dimension of hidden layers $d$ to 256 to explore the impact of several vital hyperparameters (Fig. \ref{robustness}). The performance under different hyperparameters is consistent, revealing the robustness of our method. We observe that the performance of the RCsearcher improves as the dimension of hidden layers increases. We hypothesise that the higher dimension allows the model to contain more parameters, thus improving experimental results.

\section{Conclusion and Future Work}
We present RCsearcher, a unified framework for single and multiple reaction center identification that combines the advantages of the graph neural network and deep reinforcement learning. The critical insight in this framework is that the single or multiple reaction center must be a node-induced subgraph of the molecular product graph. Comprehensive experiments demonstrate that RCsearcher consistently outperforms other baselines for the reaction centre identification task and can extrapolate the reaction center patterns that have not appeared in the training set. Ablation experiments verify the effectiveness of individual components. However, RCsearche is still limited to the reaction center of the single connected branch. In the future, we will not only take advantage of the one-hop constraint but also improve the current mechanism so that the model can generalize to the reaction center of multi-connected branches.

\newpage
\appendix
\onecolumn
\section{Dataset information}
\begin{figure*}[h]
	\centering
	\includegraphics[width=1\linewidth]{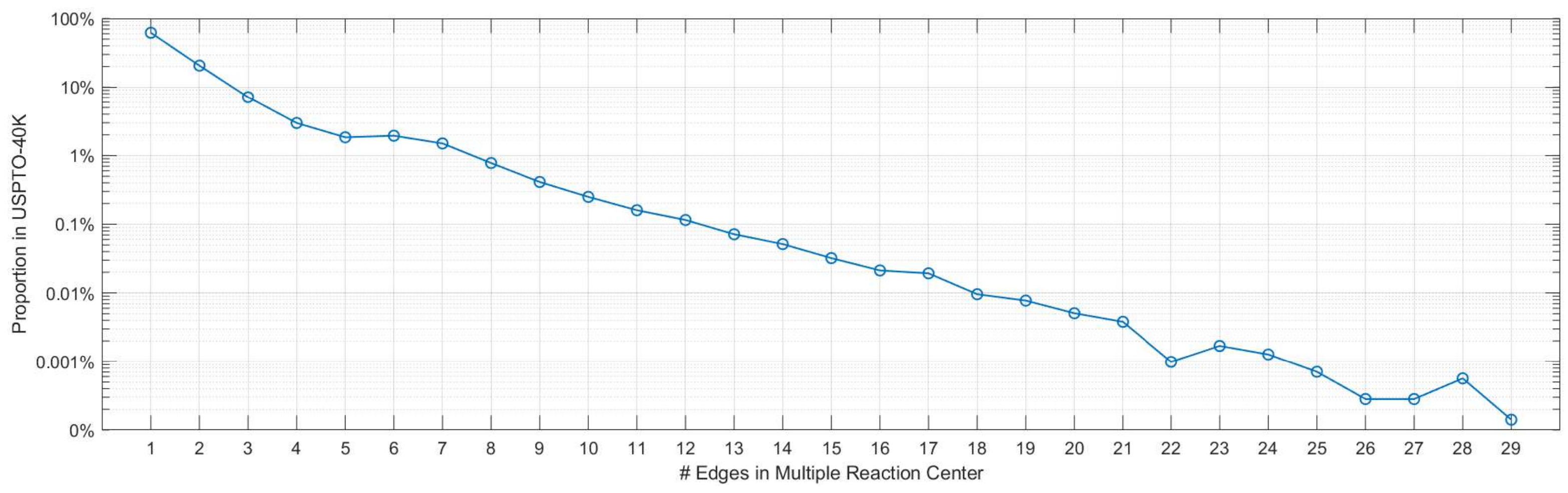}
	\caption{The detailed distribution of the USPTO-40k.}
	\label{Dataset}
\end{figure*}

We adopt stratified random sampling for USPTO-full and take out 40k samples, ensuring that the data distribution is close enough to USPTO-full. More detailed information about the distribution of the dataset can be found in Figure \ref{Dataset}.

\section{Atom and bond features}
In this paper, initial features of atom and bond can be found in Table \ref{atom} and Table \ref{bond}.

\begin{table*}[h]
\centering
\caption{Atom Features used in EGAT. All features are one-hot encoding.}
\label{atom}
\begin{tabular}{lcr}
\hline
\multicolumn{1}{c}{Feature} & Description                                    & \multicolumn{1}{c}{Size} \\ \hline
Atom type                   & Type of an atom by atomic number.              & 100                      \\
Total degree                & Degree of an atom including Hs.                & 6                        \\
Explicit valence            & Explicit valence of an atom.                   & 6                        \\
Implicit valence            & Explicit valence of an atom.                   & 6                        \\
Hybridization               & sp, sp2, sp3, sp3d, or sp3d2.                  & 5                        \\
\# Hs                       & Number of bonded Hydrogen atom.                & 5                        \\
Formal charge               & Integer electronic charge assigned to atom.    & 5                        \\
Aromaticity                 & Whether an atom is part of an aromatic system. & 1                        \\
In ring                 & Whether an atom is in ring                     & 1                        \\ \hline
\end{tabular}
\end{table*}

\begin{table*}[h]
\centering
\caption{Bond features used in EGAT. All features are one-hot encoding.}
\label{bond}
\begin{tabular}{lcr}
\hline
\multicolumn{1}{c}{Feature} & Description                          & \multicolumn{1}{c}{Size} \\ \hline
Bond type                   & Single, double, triple, or aromatic. & 4                        \\
Conjugation                 & Whether the bond is conjugated.      & 1                        \\
In ring                     & Whether the bond is part of a ring.  & 1                        \\
Stereo                      & None, any, E/Z or cis/trans.         & 6                        \\
Direction                   & The direction of the bond.             & 3                        \\ \hline
\end{tabular}
\end{table*}

\section{Hyperparameters for baselines}
For fair comparison, we set the number of layers, size of hidden dimensions and the number of attention heads in three GNN-based baselines to 4, 156, and 4 respectively. In Seq2Seq-based baseline, we set the number of layers, size of hidden dimensions and the number of attention heads in Transformer to 4, 2048, 8 respectively.

\newpage
\twocolumn[]

\bibliographystyle{icml2023}
\bibliography{re}

\end{document}